\documentclass[10pt,twocolumn,letterpaper]{article}

\usepackage[pagenumbers]{cvpr} 

\usepackage{graphicx}
\usepackage{amsmath}
\usepackage{amssymb}
\usepackage{booktabs}

\usepackage{algorithm}
\usepackage{algorithmic}
\usepackage{enumitem}
\usepackage{multirow}
\usepackage{wrapfig}
\usepackage{epsfig}
\usepackage{color}
\usepackage[table]{xcolor}
\usepackage{epsfig}
\usepackage{graphicx}
\usepackage{enumitem}

\def\cvprPaperID{4} 
\def\confName{CVPR}
\def\confYear{2023}

\begin{document}

\title{Evaluating Loss Functions and Learning Data Pre-Processing for Climate Downscaling Deep Learning Models}

\author{Xingying Huang\\
National Center for Atmospheric Research\\
Boulder, CO 80305\\
{\tt\small xyhuang@ucar.edu}
}
\maketitle

\begin{abstract}
Deep learning models have gained popularity in climate science, following their success in computer vision and other domains. For instance, researchers are increasingly employing deep learning techniques for downscaling climate data, drawing inspiration from image super-resolution models. However, there are notable differences between image data and climate data. While image data typically falls within a specific range (e.g., [0, 255]) and exhibits a relatively uniform or normal distribution, climate data can possess arbitrary value ranges and highly uneven distributions, such as precipitation data. This non-uniform distribution presents challenges when attempting to directly apply existing computer vision models to climate science tasks. Few studies have addressed this issue thus far. In this study, we explore the effects of loss functions and non-linear data pre-processing methods for deep learning models in the context of climate downscaling. We employ a climate downscaling experiment as an example to evaluate these factors. Our findings reveal that L1 loss and L2 loss perform similarly on some more balanced data like temperature data while for some imbalanced data like precipitation data, L2 loss performs significantly better than L1 loss. Additionally, we propose an approach to automatically learn the non-linear pre-processing function, which further enhances model accuracy and achieves the best results.

\end{abstract}

\section{Introduction}
\label{sec:intro}

In recent years, deep learning models have gained popularity in climate science \cite{kurth2018exascale,rolnick2022tackling,tran2019downscaling,ahmed2020multi} 
For instance, the area of climate downscaling also sees some preliminary applications \cite{vandal2017deepsd,rodrigues2018deepdownscale,pan2019improving,addison2022machine,gonzalez2023using,annau2023algorithmic} 
In the realm of computer vision tasks, images typically have pixel values ranging from 0 to 255, and these values tend to follow a normal distribution. Consequently, a common practice in deep learning models for computer vision involves pre-processing the data by subtracting the mean and dividing it by the standard deviation of the training dataset, as outlined in Krizhevsky et al.~\cite{krizhevsky2017imagenet}. The rationale behind this approach is to normalize the input data, aiming for a mean of approximately 0 and a variance of 1. Such normalization facilitates the optimization of deep learning models given this data distribution. Regrettably, climate data differs from image data in important ways. Climate data exhibits arbitrary value ranges and highly uneven distributions, making it challenging to apply the same data pre-processing techniques as those used in image analysis. For instance, precipitation data is typically non-continuous, imbalanced, and characterized by an extremely wide dynamic range. Conversely, other climate data, such as temperature data, tends to have a more balanced distribution and a narrower dynamic range. These disparities in data distribution hinder the direct adoption of existing deep learning models from computer vision to climate-related tasks, necessitating modifications in model design. These adjustments include selecting appropriate loss functions and employing advanced data pre-processing methods, among other strategies.

In this study, we leverage deep climate downscaling as a case study and conduct experiments on both precipitation and temperature data. Our primary objective is to investigate the impact of employing different loss functions, specifically L1 loss and L2 loss. Furthermore, we assess the efficacy of non-linear preprocessing techniques and determine whether they contribute to improved results. Lastly, we propose a learnable non-linear preprocessing scheme that enables the model to automatically learn the optimal data preprocessing approach. Quantitative and Qualitative results show that: 

\setlist{nolistsep}
\begin{itemize}[noitemsep]
\item L1 loss and L2 loss perform similarly for temperature data while L2 loss performs significantly better than L2 loss for precipitation data.

\item Predefined non-linear data preprocessing is not always helpful for climate data. For example, gamma correction with fixed $\gamma=2.2$ improves the precipitation downscaling model but worse the quality of the temperature downscaling model.

\item Learning the non-linear data preprocessing automatically improves the downscaling quality for both the precipitation downscaling model and temperature downscaling model.

\end{itemize}

\section{Related Work}

With the increasing demand for high-resolution climate data across emerging climate studies and real-world needs~\cite{giorgi2009addressing,stocker2014summary,roberts2018benefits},
rapidly growing efforts have focused on developing methods and techniques to retrieve fine-scale details from coarse-resolution sources either from reanalysis or simulations~\cite{wood2004hydrologic,maraun2010precipitation,giorgi2015regional}
Existing downscaling methods mainly include but not limited to traditional dynamical downscaling (using either regional climate models, variable-resolution global climate modeling, or high-resolution global climate models), and empirical statistical downscaling (either linear or nonlinear), attributing with unique strengths and also limitations~\cite{huang2016evaluation}.

Deep learning has also been used for climate data downscaling. Vandal et al.~\cite{vandal2017deepsd} presented a DeepSD framework composed of three convolutional layers. And Rodrigues et al.~\cite{rodrigues2018deepdownscale} explored climate downscaling using several convolution layers and locality-specific layers. Further, Pan et al~\cite{pan2019improving} used convolutional layers and fully connected layers, with every entry in the input being connected to every entry in the output regardless of their locations, to predict per-grid point value. Recently more advanced generative models have also been used for climate data downscaling. Annau et al. \cite{annau2023algorithmic} designed a conditional Generative Adversarial Network to generate high resolution climate data output conditioned on low resolution climate data input. Inspired by the success of diffusion models~\cite{ho2020denoising} in image synthesis, Addison et al.~\cite{addison2022machine} extended image super resolution diffusion model for climate data downscaling task.

\section{Loss Function}
Researchers have explored a range of loss functions for generative models, which include L1 loss, L2 loss, perceptual loss~\cite{simonyan2014very}, GAN loss~\cite{goodfellow2020generative}, KL-Divergence loss~\cite{kingma2013auto} etc. In this study, our primary focus is on evaluating the effectiveness of L1 loss and L2 loss for climate data. These two loss functions are often considered foundational and serve as the basis for other advanced loss functions. Additionally, they are sometimes used in conjunction with more sophisticated loss functions. The L1 loss and L2 loss functions are defined as the following:

\begin{equation}
    \mathcal{L}_{1} =  \sum_{i \in N}{\frac{\|\mathbf{I}_{pred}^{i} - \mathbf{I}_{gt}^{i}\|}{N}}, \mathcal{L}_{2} =  \sum_{i \in N}{\frac{(\mathbf{I}_{pred}^{i} - \mathbf{I}_{gt}^{i})^2}{N}}
\end{equation}

where $\|*\|$ represents the absolute difference, $\mathbf{I}_{pred}$ and $\mathbf{I}_{gt}$ denotes the prediction and ground truth respectively and $N$ is the total number of items. 

\section{Learning Non-Linear Pre-Processing}
In deep learning, data normalization is a commonly employed technique to preprocess input data for deep learning models. For instance, in the ImageNet classification model\cite{krizhevsky2017imagenet}, the input image will first go through the process of subtracting the mean and dividing the standard deviation of the whole set of training images, to roughly normalize the input data to normal distribution. The hypothesis is that deep learning models are easier to train if the input data is with simpler distribution. 

Climate data tends to have a significantly more intricate distribution compared to image data, and this can be attributed to several factors. Firstly, unlike image data where values typically range from 0 to 255, climate data does not have predefined boundaries for its value range. For instance, precipitation data can reach exceptionally high values  while the minimal value may be some small decimals around zero. 2). The climate data values often exhibit a high degree of imbalance in their distribution.

\subsection{Gamma Correction}
\label{sec:gamma_correct}
Gamma correction is a commonly used tool to perform nonlinear mapping on the pixel values of images~\cite{gamma_wiki}. The gamma correction function is defined as Equation \ref{eq:gamma}. 

\begin{equation}
    f(x) = x^{1/\gamma}
    \label{eq:gamma}
\end{equation}

Since Equation~\ref{eq:gamma} only applies for non-negative input values while climate data may have negative values like temperature, we extend Equation~\ref{eq:gamma} to be Equation~\ref{eq:gamma_ext} to handle negative values.

\begin{equation}
    f(x) = sign(x) * |x|^{1/\gamma}
    \label{eq:gamma_ext}
\end{equation}

where $sign(x) = \frac{x}{|x|}$ denotes the sign of $x$. 

When gamma value $\gamma$ is $>1$, the function becomes a compressive power-law nonlinear mapping process and this gamma correction is called \textbf{gamma compression}. When $\gamma$ is $<1$, the function becomes an expansive power-law nonlinear mapping process and this gamma correction is called \textbf{gamma expansion}. In graphics, the $\gamma$ is usually set to be $2.2$, where the goal is to compress the high dynamic range values to the low dynamic range pixel values (e.g. [0, 255]) that can be displayed in the monitor and perceived by human eyes properly. 

\subsection{Learning Gamma Value Automatically}

In graphics, the most common use case of gamma correction is to map the saturated synthesized pixel values to low dynamic range color values (e.g. [0, 255]). Thus, $\gamma$ is mostly set to be $>1$ (empirically 2.2). However, for climate data, it is not always the case. The climate data values can be either saturated or under-saturated. Precipitation data may have many saturated values where some regions have floods while other regions have zero precipitation. For such cases, $\gamma$ needs to be $>1$. In contrast, temperature data may be under-saturated when all the study regions have very similar and stable temperatures, where $\gamma<1$ will be the ideal setting.

Thus, we further extend Equation~\ref{eq:gamma_ext} to letting the model learn the gamma value $\gamma$ automatically. $\gamma$ will be trained together with other parameters in the model. The goal is to learn the $\gamma$ value so that it can adapt to different types of climate automatically.

\section{Experiment}

To evaluate the effects of loss functions and non-linear pre-processing for climate data, we choose climate data downscaling task for our experiments. Specifically, we use a UNet model to downscale the precipitation and temperature data and compare the results obtained using different loss functions and non-linear pre-processing settings. 

\subsection{Precipitation and Temperature Data}
We use daily precipitation data and temperature data for the downscaling experiments. Temperature data represents the data type with balanced distribution while precipitation data represents the extremely imbalanced data. Most of the precipitation data values may be close to zero while the large precipitation can be extremely big.
Daily data is targeted covering the whole western US from 1981 to 2010. The goal is to downscale the coarse-resolution reanalysis input (here, using ERA-interim, ~81 km) to 4 km (also the resolution of the ground-truth) for near-surface (2 m) temperature (T2) and precipitation (Pr) for each day. 
In detail, ERA-interim, a widely-used reanalysis dataset~\cite{dee2011era}, is chosen as the coarse-resolution input. A well-received high-quality gridded observational dataset, PRISM (Parameter-elevation Regressions on Independent Slopes Model~\cite{daly2008physiographically}), is applied as the ground truth for training purposes. 

Supporting datasets have been used together with the input for network training purposes, including elevations from both input and ground-truth sources at different native grid resolutions for both temperature and precipitation downscaling. Due to the discontinuity and complexity of the precipitation field, additional supporting datasets have been used in addition to the elevations, including zonal and meridional winds (U and V), relative humidity (RH), and specific humidity (Q) from the coarse-resolution input data (i.e. ERA-interim) at 850 hPa vertical level. 
For analysis convenience, all the datasets have been regridded to 4 km using the bilinear method.

\subsection{Network Design}

UNet design is employed for the experiment. The encoder contains 3 groups. Each group contains a stride=2 convolution block and a stride=1 convolution block, with each convolution block consisting of a convolution layer, a batch normalization layer, and a ReLU layer. The decoder also has 3 groups. Each group contains an upsampling block and stride=1 convolution block. The upsampling block includes a nearest upsampling layer, a stride=1 convolution layer, a batch normalization layer, and a ReLU layer. Batch normalization and ReLU layers are not used in the last convolution block. We empirically find that using two skip link connections in the middle part of the network performs the best compared with using one or three skip connections and is used for all the experiments. The total number of parameters is 7.5 million. All the models are trained for 800 epochs on 8 V100 GPUs.

\subsection{Comparisons}

\begin{table} 
\centering
\begin{tabular}{|l|c|c|}
\hline
Method & avg ABS diff. ($\downarrow$) & avg MSE ($\downarrow$)\\
\hline
L1 & 0.8846 & 14.3256 \\
L2 & 0.9714 & 11.8747 \\
\hline
L1 + NL2.2 & 0.8704 & 12.9761 \\
L1 + Learn & 0.8717 & 12.8292 \\
\hline
L2 + NL2.2 & 0.9676 & 12.3032 \\
L2 + Learn & 0.9587 & 12.1436 \\
\hline
\end{tabular}
\caption{Comparisons with various methods for precipitation prediction. avg ABS diff. and avg MSE represent the average absolute difference and mean square errors between ground truth and prediction.}
\label{tab:pr}
\vspace{-0.3cm}
\end{table}


\begin{table} 
\centering
\begin{tabular}{|l|c|c|}
\hline
Method & avg ABS diff. ($\downarrow$) & avg MSE ($\downarrow$)\\
\hline
L1 & 0.9200 & 2.157 \\
L2 & 0.9214 & 2.130 \\
\hline
L1 + NL2.2 & 0.9151 & 2.136 \\
L1 + Learn & 0.8833 & 1.988 \\
\hline
L2 + NL2.2 & 1.2694 & 3.968 \\
L2 + Learn & 0.8993 & 2.033 \\
\hline
\end{tabular}
\caption{Comparisons with various settings for temperature prediction. avg ABS diff. and avg MSE represent the average absolute difference and mean square errors between ground truth and prediction.}
\label{tab:t2}
\vspace{-0.3cm}
\end{table}

\begin{figure*}
    \centering
    {\includegraphics[width=1.4\columnwidth,trim={200 1000 200 970},clip=true]{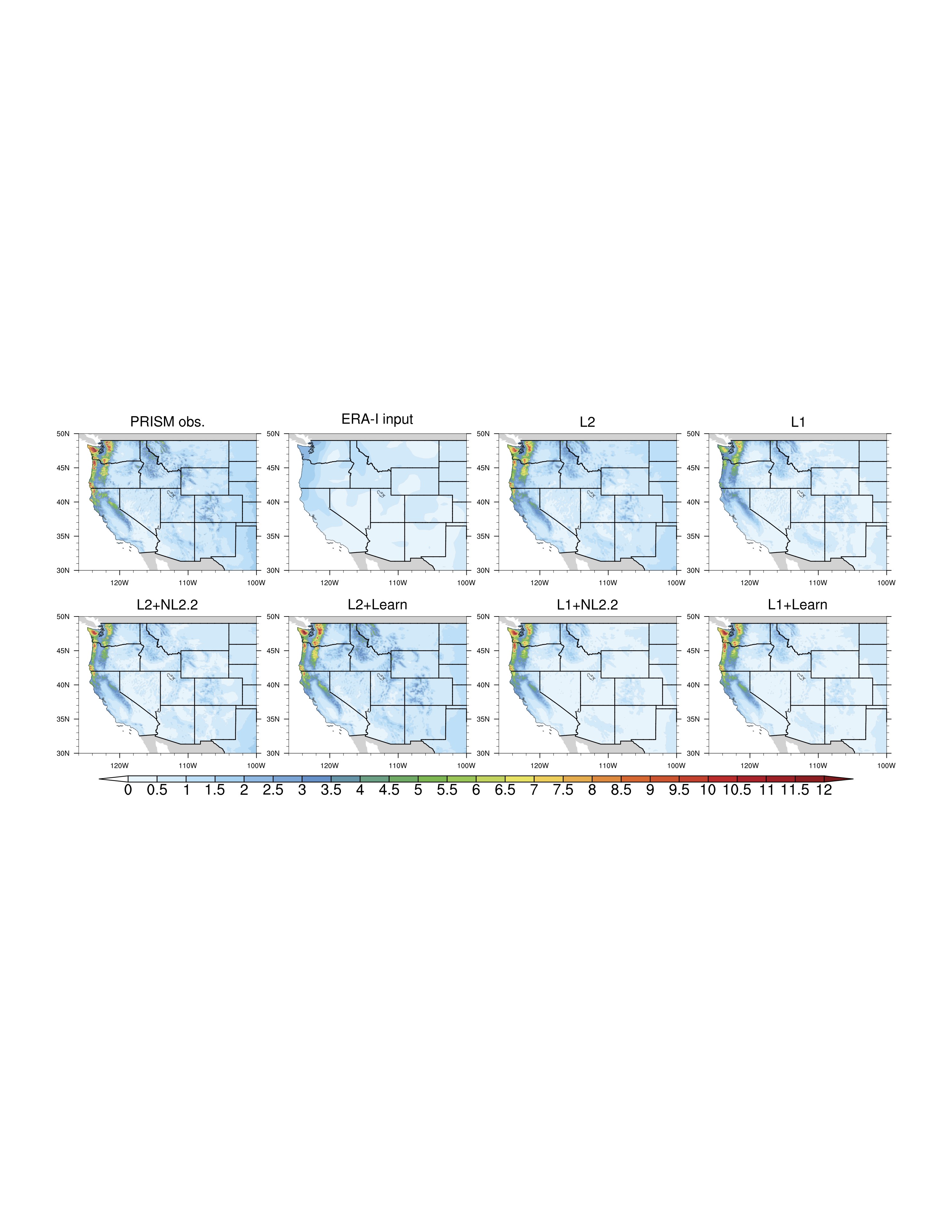}}
    {\includegraphics[width=1.4\columnwidth,trim={200 1000 200 968},clip=true]{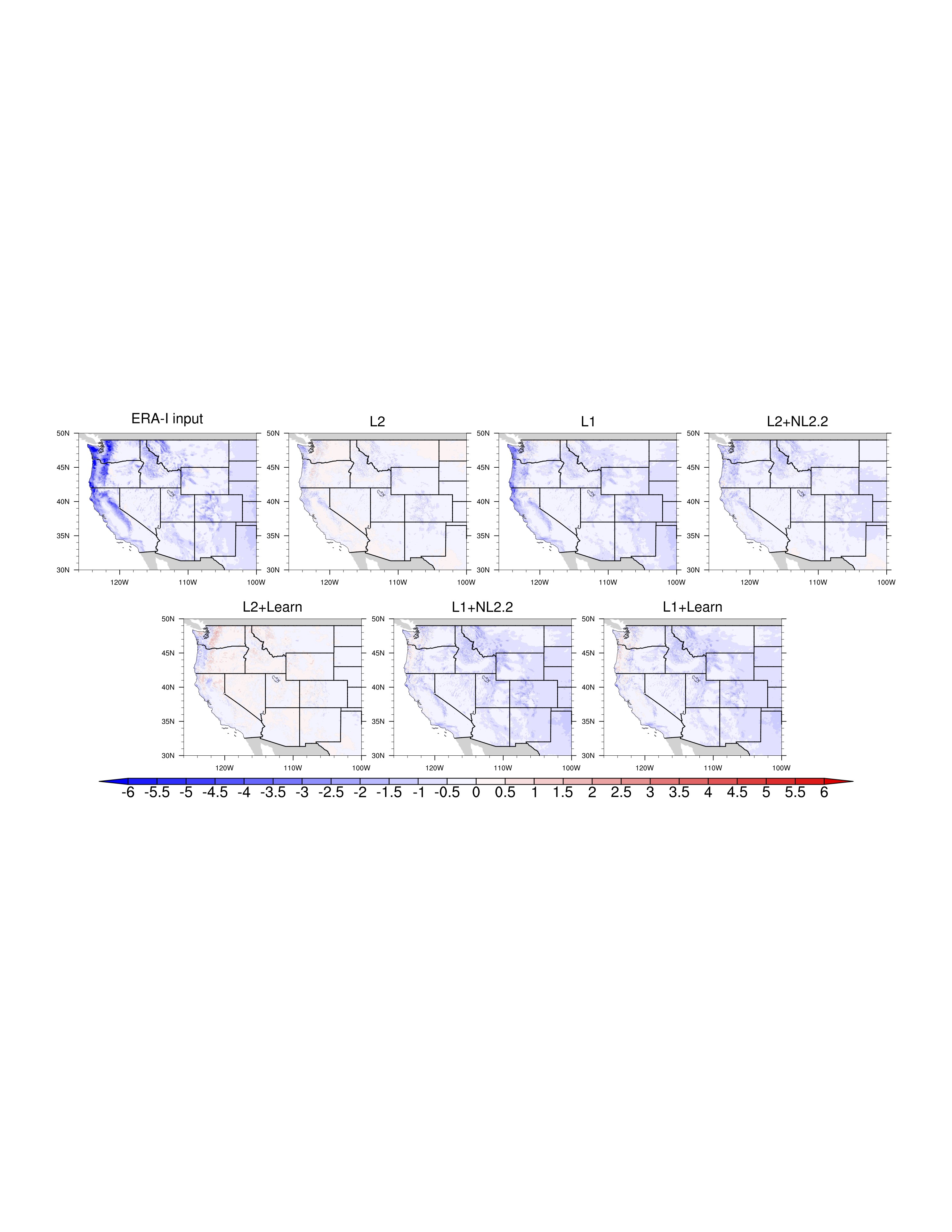}}
    \caption{Visualization of precipitation downscaling results. \textbf{Top two rows} show the yearly mean precipitation over 2001-2010. \textit{PRISM obs.} represents the high-resolution ground truth. \textit{ERA-I input} denotes the coarse resolution input. \textit{L2} and \textit{L1} represents the results are obtained using L2 and L1 losses respectively. $NL2.2$ represents the result is obtained using gamma correction with fixed $\gamma=2.2$ while $Learn$ uses learnable $\gamma$. \textbf{Bottom two rows} shows the difference between the ground truth and outputs from different settings.}
    \label{fig:vis_pr}
\end{figure*}

\begin{figure*}
    \centering
    \includegraphics[width=1.45\columnwidth,trim={200 1000 200 970},clip=true]{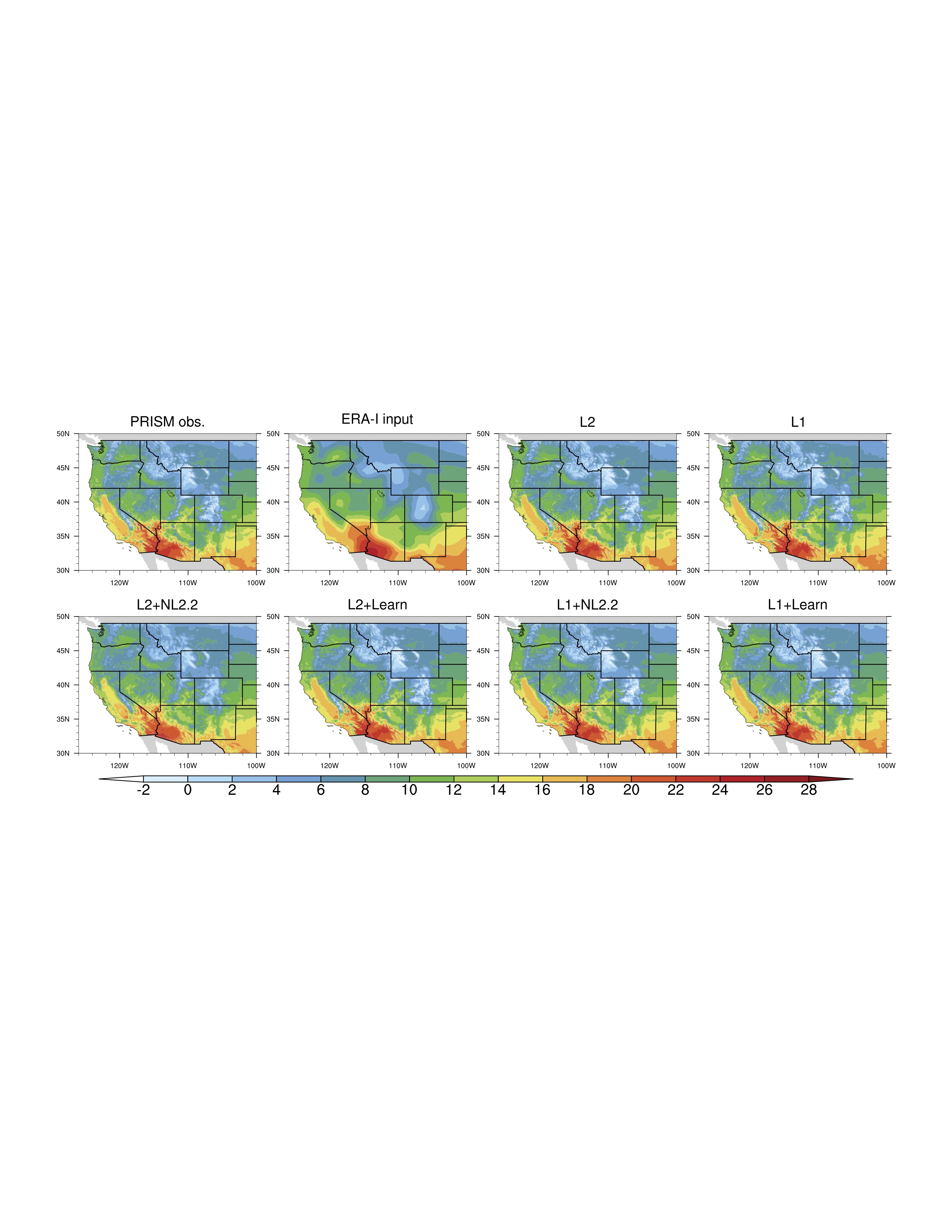}
    \includegraphics[width=1.45\columnwidth,trim={200 1000 200 968},clip=true]{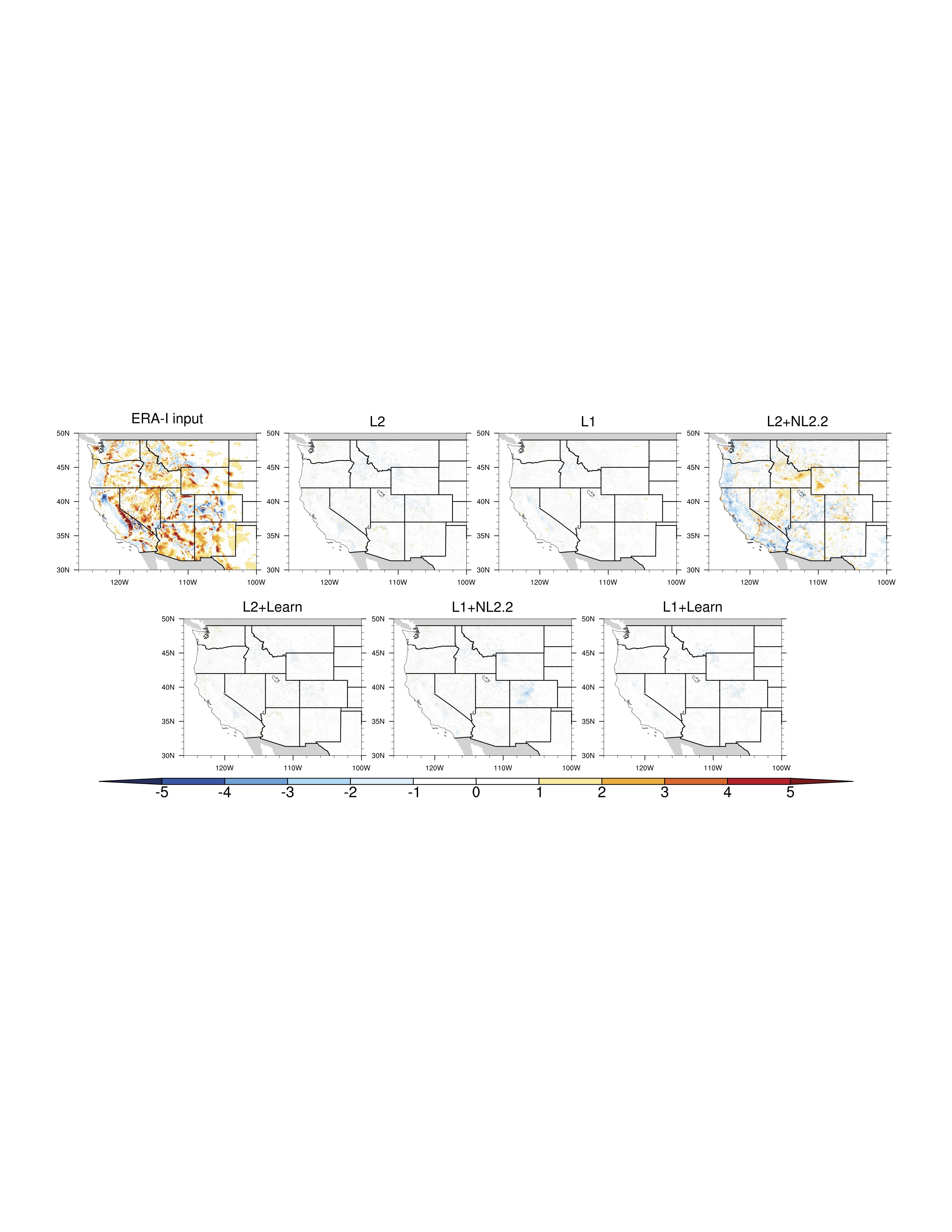}
    \caption{Visualization of temperature downscaling results. \textbf{Top two rows} show the yearly mean precipitation over 2001-2010. \textit{PRISM obs.} represents the high-resolution ground truth. \textit{ERA-I input} denotes the coarse resolution input. \textit{L2} and \textit{L1} represents the results are obtained using L2 and L1 losses respectively. $NL2.2$ represents the result is obtained using gamma correction with fixed $\gamma=2.2$ while $Learn$ uses learnable $\gamma$. \textbf{Bottom two rows} shows the difference between the ground truth and outputs from different settings.}
    \label{fig:vis_t2}
\end{figure*}

Figure~\ref{fig:vis_pr} and Figure~\ref{fig:vis_t2} show the visualization of precipitation data downscaling results and temperature data downscaling results respectively. Quantitative comparisons are shown in Table~\ref{tab:pr} and Table~\ref{tab:t2}. For all these tables and figures, the specific settings are the following:

\setlist{nolistsep}
\begin{itemize}[noitemsep]
\item \textbf{PRISM obs.}: ground truth high-resolution observational data.
\item \textbf{EAR-I input}: coarse resolution input data.
\item \textbf{L2}: result from model trained with L2 loss.
\item \textbf{L1}: result from model trained with L1 loss.
\item \textbf{L2+NL2.2}: result from model trained with L2 loss and input data is pre-processed with gamma correction where $\gamma$ is fixed to be $2.2$.
\item \textbf{L2+Learn}: result from model trained with L2 loss and input data is pre-processed with gamma correction where $\gamma$ is learnable.
\item \textbf{L1+NL2.2}: result from model trained with L1 loss and input data is pre-processed with gamma correction where $\gamma$ is fixed to be $2.2$.
\item \textbf{L1+Learn}: result from model trained with L1 loss and input data is pre-processed with gamma correction where $\gamma$ is learnable.
\end{itemize}

\subsubsection{L1 loss vs. L2 loss}
If judging only from quantitative comparisons from the first two rows of Table~\ref{tab:pr} and Table~\ref{tab:t2}, it is impossible to conclude which of \textbf{L1} loss and \textbf{L2} loss is better over the other one. Either loss function will get better testing scores on the metric that it is optimized for during the training. Namely, \textbf{L1} loss would get a lower average absolute difference but a higher average mean square difference on testing set.

For precipitation data, Figure~\ref{fig:vis_pr} shows that qualitatively \textbf{L2} performs significantly better than \textbf{L1}. The spatial patterns are much better represented in the result obtained using \textbf{L2}, especially over the west US regions with heavy precipitation. The model with \textbf{L1} loss often underestimate the precipitation values for heavy precipitation regions. For temperature data, \textbf{L1} perform similarly or slightly better compared with \textbf{L2} shown in the 3rd row of Figure~\ref{fig:vis_t2}.

\subsubsection{Non-linear pre-processing}
\label{sec:fix_gamma}

For precipitation data, using gamma correction with fixed $\gamma=2.2$ to pre-process the input data usually obtain results than not using it. In both Table~\ref{tab:pr} and Figure~\ref{fig:vis_pr}, \textbf{L1+NL2.2} performs better than \textbf{L1} and \textbf{L2+NL2.2} performs better than \textbf{L2}. However, for temperature data, gamma correction with fixed $\gamma=2.2$ may lead to even worse results. In Table~\ref{tab:t2}, \textbf{L2+NL2.2} performs much worse than \textbf{L2}.

As discussed in Section~\ref{sec:gamma_correct}, $\gamma$ is $>1$ corresponds compressive preprocessing. This indicates that applying compressive preprocessing aids in improving the convergence of the precipitation downscaling model. Conversely, compressive preprocessing has a detrimental effect on the temperature downscaling model. We propose that this discrepancy stems from the inherent characteristics of the data. Precipitation data typically exhibits a high dynamic range where compressive preprocessing is helpful, whereas temperature data tends to have a lower dynamic range. Compressive preprocessing further reduces the dynamic range of the temperature data, making it more challenging to optimize the model within such a constrained range.

It implies that some data needs compressive preprocessing (with $\gamma>1$) while some data may need expansive preprocessing (with $\gamma<1$). Thus, we design experiments to learn $\gamma$ automatically. Table~\ref{tab:t2} shows that the \textbf{L*+Learn} always perform better than both \textbf{L*} and \textbf{L*+NL2.2}. Table~\ref{tab:pr} shows \textbf{L*+Learn} performs better than \textbf{L*+NL2.2}. Figure~\ref{fig:vis_pr} and Figure~\ref{fig:vis_t2} also show that \textbf{L*+Learn} get better visualization results than either \textbf{L*} or \textbf{L*+NL2.2}.

\section{Summary}
In this work, we evaluate different loss functions and data preprocessing strategies for deep climate downscaling models. Experiments show that L2 loss performs significantly better than L2 loss for precipitation data though they perform similarly for temperature data. Predefined non-linear preprocessing like gamma correction with $\gamma=2.2$ improves precipitation downscaling but worsens temperature downscaling. Finally, we propose to learn the non-linear data preprocessing automatically which improves both precipitation and temperature downscaling.


{\small
\bibliographystyle{ieee_fullname}
\bibliography{egbib}
}

\end{document}